\documentclass[11pt,a4paper]{article}
\usepackage[dvipsnames]{xcolor}
\usepackage{times,latexsym}
\usepackage{url}
\usepackage[T1]{fontenc}
\usepackage{linguex}
\usepackage{booktabs}
\usepackage{graphicx}
\usepackage{amssymb}
\usepackage{amsmath}
\usepackage{hyperref}
\usepackage[all]{hypcap}
\usepackage{bibentry}

\usepackage[acceptedWithA]{tacl2021v1}

\usepackage{xspace,mfirstuc,tabulary}

\newif\iftaclinstructions
\taclinstructionsfalse 
\iftaclinstructions
\renewcommand{\confidential}{}
\renewcommand{\anonsubtext}{(No author info supplied here, for consistency with
TACL-submission anonymization requirements)}
\newcommand{\instr}
\fi

\iftaclpubformat 

\else

\fi

\definecolor{beaublue}{rgb}{0.74, 0.83, 0.9}
\definecolor{babypink}{rgb}{0.96, 0.76, 0.76}

\title{Uncontrolled Lexical Exposure Leads to Overestimation of\\Compositional Generalization in Pretrained Models}

\author{
    Najoung Kim$^*$ \\
    Boston University \\
    \texttt{najoung@bu.edu}
\And
    Tal Linzen \\
    New York University \\
    \texttt{linzen@nyu.edu}
\And
    Paul Smolensky \\
    Johns Hopkins University \\
    Microsoft Research, Redmond \\
    \texttt{smolensky@jhu.edu}
}

\date{}

\begin{document}
\maketitle
\begin{abstract}
Human linguistic capacity is often characterized by compositionality and the generalization it enables---human learners can produce and comprehend novel complex expressions by composing known parts. Several benchmarks exploit distributional control across training and test to gauge compositional generalization, where certain lexical items only occur in limited contexts during training. While recent work using these benchmarks suggests that pretrained models achieve impressive generalization performance, we argue that exposure to pretraining data may break the aforementioned distributional control. Using the COGS benchmark of \citet{kim2020cogs}, we test two modified evaluation setups that control for this issue: (1) substituting context-controlled lexical items with novel character sequences, and (2) substituting them with special tokens represented by novel embeddings. We find that both of these setups lead to lower generalization performance in T5 \citep{raffel2020exploring}, suggesting that previously reported results have been overestimated due to uncontrolled lexical exposure during pretraining. The performance degradation is more extreme with novel embeddings, and the degradation increases with the amount of pretraining data, highlighting an interesting case of inverse scaling.
\end{abstract}

{\let\thefootnote\relax\footnotetext{$^{*}$Work partially done at New York University and Johns Hopkins University.}}

\section{An Issue in Testing Pretrained Models for Compositional Generalization}
\label{sec:issue}
Compositional generalization, the ability to produce and comprehend novel complex expressions by composing known parts, has been considered a key property of human cognitive and linguistic capacity \citep{frege1923compound,fodor1988connectionism,smolensky1991constituent,lake2017building}. This ability is also desirable for coverage of long-tail phenomena in practical applications such as semantic parsing and question-answering \citep{finegan-dollak2018improving,liu2021challenges}. Compositional generalization has traditionally been considered a challenge for neural network models \citep{fodor1988connectionism,hadley1994systematicity,phillips1998feedforward,van2004lack}, and in this context, benchmarks such as SCAN \citep{lake2018generalization}, COGS \citep{kim2020cogs}, CFQ \citep{keysers2020measuring} and SyGNS \citep{yanaka2021sygns} have recently been proposed to evaluate models' generalization capacity. These benchmarks motivated active efforts for improvement and analysis \citep[\textit{i.a.}]{liu2021learning,jiang2021inducing,ontanon2022making,bogin2022unobserved,jambor2022lagr}. Among such work, some report that models pretrained on context reconstruction (typically, ``language modeling'') such as T5 \citep{raffel2020exploring}, mT5 \citep{xue2021mt5}, CodeT5 \citep{wang2021codet5} and pretrained convolutional sequence-to-sequence (seq2seq) networks achieve high generalization accuracy on SCAN and COGS \citep{shaw2021compositional,tay2021pretrained,orhan2021compositional}.

A core property of many compositional generalization benchmarks is the existence of distributional mismatches between training and generalization sets that can be overcome by composing parts of the training examples in an appropriate way. For example, the COGS training set contains a sentence with the noun \textit{hedgehog} appearing as a part of a subject noun phrase (e.g., \textit{The hedgehog saw the cat}), and the generalization set contains examples with \textit{hedgehog} as a part of an object noun phrase (e.g., \textit{The cat saw the hedgehog}). Importantly, sentences with \textit{hedgehog} as a part of an object noun phrase are absent from the training set, which creates a distributional mismatch between training and generalization. The primitive generalization split in SCAN exploits a similar idea: complex examples containing certain primitives (e.g., \textit{jump}) are withheld from the training set. Furthermore, there is often a limited exposure component to the compositional generalization benchmarks: there is only a limited number of examples that expose the models to the \textbf{context-controlled} lexical items like \textit{hedgehog}. COGS limits the number of exposure examples to 1, and SCAN's primitive splits limit exposure to a single example at the type-level (a single exposure example comprises 10\% of the training set). This design is analogous to, or sometimes explicitly takes motivation from, human subject experiments that use nonce words to test generalization. Importantly, the critical assumption of such experiments is that subjects would not have encountered the nonce words prior to the experiment, so that their exposure to those words can be completely controlled.

One property that the aformentioned benchmarks share is that the context-controlled lexical items are real words of English like \textit{hedgehog} and \textit{jump}. This poses no issue when a model is trained \textit{only} on the training sets of these benchmark datasets.  However, if a model is trained on any data additional to the benchmark's training set, it is no longer guaranteed that the distributional control intended by the benchmark still holds, unless one inspects the additional data fully to ensure that there are no examples of the type intended to be withheld from training. To summarize, additional training data can violate the assumptions of these benchmarks in the following ways:

\begin{itemize}
    \item Lexical items that are intended to be withheld from specific contexts (\textbf{context-controlled}) are observed in such contexts during pre/auxiliary training.
    \item Lexical items that are intended to have limited numbers of exposures are observed more frequently during pre/auxiliary training.
\end{itemize}

\noindent Large pretrained models almost certainly fall under these scenarios. For instance, it is unlikely that there is no occurrence of \textit{hedgehog} as a part of an object noun phrase, no occurrence of \textit{jump} as a part of a complex expression, or no more than 1 occurrence of these context-controlled items in the Colossal Clean Crawled Corpus (C4; \citealt{raffel2020exploring}), the pretraining corpus of the T5 models commonly used in the literature to tackle these generalization benchmarks.

We propose two modified evaluation setups that control for this issue: (1) using  as context-controlled lexical items character sequences that do not occur in the pretraining data (tokenized and embedded via the model's existing tokenization process), or (2) directly using a single novel embedding to initialize each context-controlled lexical item. In the new experiments reported here, under both setups, the pretrained T5 models performed strictly worse than their performance on the unmodified version of the COGS dataset. We refer to this performance gap as \textit{overestimation due to uncontrolled lexical exposure}, shortened as simply \textbf{overestimation}. Under setup (1), the overestimation was between 14 and 19 percentage points. Under setup (2), the overestimation was much larger at around 51 percentage points. Under setup (2), the performance degradation was inversely correlated with the amount of language modeling pretraining data: pretrained models performed \textit{worse} than randomly initialized models of the same architecture.

Overall, our results support the conclusion that previously reported generalization performance of pretrained models has been substantially overestimated, and furthermore highlight the surprising sensitivity of the models' generalization behavior to the choice of the type of the context-controlled lexical items.\footnote{Code and data available at: \url{https://github.com/najoungkim/cogs-with-pretraining}}

\section{Proposed Modifications}
\label{sec:modification}
In light of the issue raised in Section~\ref{sec:issue}, we propose two modifications to the compositional generalization dataset and evaluation setup of \citet{kim2020cogs} that guarantee the intended distributional control across training and generalization. Although we focus on COGS as a case study here, similar modification should be applicable to SCAN or other tests that gauge generalization to novel contextual usages of lexical items. 

In the original COGS dataset, real English words were used as context-controlled lexical items (e.g., \textit{hedgehog}, \textit{cockroach}).\footnote{Note that there are broadly two types of generalization in COGS: lexical (novel combination of a familiar lexical item and a familiar linguistic structure) and structural (novel structures). We focus on lexical generalization, for which a benefit of pretraining has been claimed in the literature, but see Appendix~\ref{app:structural-generalization} for more discussion about structural generalization.} Our proposal is to replace these with either character sequences that do not appear in the pretraining data (e.g., \textit{bahufowu}), or to replace them with special tokens (e.g., [$w_n$]\footnote{The surface forms of the replacement tokens do not matter under this setup. The only requirement is that the tokens selected do not already exist in the model's vocabulary.}) that are newly added to the vocabulary of the model being tested. Example replacements are shown in \ref{ex:modification}:

\ex. \label{ex:modification}
    \a. \textsc{Orig.:} Emma liked the \colorbox{babypink}{hedgehog}. \\$\leadsto$ *\colorbox{babypink}{hedgehog}($x_3$); like.agent($x_1$, Emma) \textsc{and} like.theme($x_1$, $x_3$)
    \vspace{0.1cm}
    \b. \textsc{Mod 1:} Emma liked the \colorbox{beaublue}{bahufowu}. \\$\leadsto$ *\colorbox{beaublue}{bahufowu}($x_3$); like.agent($x_1$, Emma) \textsc{and} like.theme($x_1$, $x_3$)
    \vspace{0.1cm}
    \c. \textsc{Mod 2:} Emma liked the \colorbox{beaublue}{[$w_0$]}. \\$\leadsto$ *\colorbox{beaublue}{[$w_0$]}($x_3$); like.agent($x_1$, Emma) \textsc{and} like.theme($x_1$, $x_3$)

\noindent We applied this substitution to both the input sentence and the output logical form, substituting each unique context-controlled item with a different novel character sequence or a special token represented by a novel embedding.

We tested two different approaches because the proposed modifications have different pros and cons, although they both provide control for lexical exposure. In the case of character sequence substitution, we need an additional step to verify that these sequences indeed do not occur in the pretraining data. This step may not always be feasible if the pretraining corpus is inaccessible. On the other hand, non-occurrence in the pretraining data is always guaranteed under the novel embeddings setup, since the replacement tokens are novel entries in the models' vocabulary, added after pretraining. However, adding new tokens requires a modification to the model, leading to additional experimental choices such as the initialization scheme as we discuss in Section~\ref{exp2:novel-embeddings}.

\subsection{Test Set for Lexical Difficulty}
\label{subsec:id-test-novel-lex}
Before we describe our main experiments, we briefly introduce another test set, which we refer to as a test set for lexical difficulty (\textsc{Test-Lex}). The original COGS dataset contains both in-distribution test examples (\textsc{Test-ID}) \ref{ex:id-test} and out-of-distribution generalization examples (\textsc{Gen}) \ref{ex:gen}. However, the original test set only contains recombinations of non-context-controlled items as in \ref{ex:id-test}. This means there are no examples like \ref{ex:id-test-novel} in \textsc{Test-ID}, where context-controlled items appear in the same type of contexts as their exposure examples in the training set \ref{ex:exposure} (e.g., different examples with \textit{hedgehog} as part of a subject noun phrase when the training set already showed \textit{hedgehog} as a part of a subject noun phrase).

\ex.
    \a. \label{ex:exposure} \textsc{Training:} The hedgehog/bahufowu\\/[$w_0$] ate the cake.\\The girl saw the donut.
    \b. \label{ex:id-test} \textsc{Test-ID:} The girl ate the cake.
    \c. \label{ex:id-test-novel} \textsc{Test-Lex:} The hedgehog/bahufowu\\/[$w_0$] ate the donut.
    \d. \label{ex:gen} \textsc{Gen (Subj-to-Obj):} The girl saw the hedgehog/bahufowu/[$w_0$].

\textsc{Test-Lex} consists of new in-distribution uses of the context-controlled lexical items ($n=12,000$). The goal is to better tease apart the difficulty of processing less familiar lexical items (i.e., novel character sequences or novel embeddings) from the difficulty of bridging the distributional gap across training and generalization through composition. Note that the latter difficulty only exists in the generalization examples \ref{ex:gen}, whereas the former exists in both the generalization \ref{ex:gen} and the \textsc{Test-Lex} \ref{ex:id-test-novel} examples.

\begin{table*}[t]
    \centering
    \resizebox{2\columnwidth}{!}{
    \begin{tabular}{ccccccc}
    \toprule    
           Length & Character distribution & Example & \colorbox{beaublue}{\underline{\textbf{Gen.}}} & Test-ID & Gen. (Lex. only) & Test-Lex \\ \midrule\midrule
           Longer & Random & \textit{rkijtgjqamjtwsmcbi} & 0.681 \small{($\pm$ 0.022)} & 0.998 & 0.786 \small{($\pm$ 0.025)} & 0.783 \small{($\pm$ 0.014)} \\
           Shorter & Random & \textit{dvalcxw} & 0.692 \small{($\pm$ 0.016)} & 0.998 & 0.798 \small{($\pm$ 0.019)} & 0.750 \small{($\pm$ 0.030)}\\
           Longer & CVCV & \textit{tayutenotipevobe} & 0.690 \small{($\pm$ 0.018)} & 0.998 & 0.795 \small{($\pm$ 0.021)} & 0.739 \small{($\pm$ 0.020)} \\
           Shorter & CVCV & \textit{bahufowu} & 0.642 \small{($\pm$ 0.020)} & 0.998 & 0.740 \small{($\pm$ 0.023)} & 0.699 \small{($\pm$ 0.047)} \\\midrule
           \multicolumn{3}{c}{No modification (replication of \citealt{orhan2021compositional})} & 0.833 & 0.998 & 0.963 & 0.973 \\
           \bottomrule
    \vspace{0.01cm}
    \end{tabular}
    }
    \caption{Generalization accuracy of T5-base trained on datasets with context-controlled lexical items replaced with sampled character sequences. \textbf{Gen.} refers to accuracy on the full generalization set comparable to performance reports in the literature. \textbf{Gen. (Lex. only)} lists the performance on the lexical generalization portion of the dataset, excluding structural generalization, for fair comparison to \textbf{Test-Lex} that only contains lexical generalization. Standard deviations over five random seeds are shown if greater than 0.01.}
    \label{table:char-sampling}
\end{table*}

\section{Experiment 1: Novel Character Sequences as Context-controlled Lexical Items}

\subsection{Character Sampling}
As discussed in Section~\ref{sec:modification}, we modified the original dataset by replacing context-controlled lexical items with novel character sequences. We sampled these sequences from the 26 lower-case ASCII alphabet characters with replacement. We furthermore varied this sampling process along two dimensions that may affect generalization: length and character distribution within the sequence (random sampling vs. alternating between consonants and vowels). For length, we either sampled shorter ([7--15) chars) or longer strings ([15--30) chars). Between the random sampling and consonant-vowel alternation sampling, the latter is likelier to yield character sequences that are closer to real lexical items of English (e.g., \textit{bahufowu}) than random sampling (e.g., \textit{dvalcxw}), in terms of transition probabilities between the characters or subsequences that comprise the sampled sequence. We crossed these two factors, length (\texttt{longer} vs. \texttt{shorter}) and character distribution (\texttt{random} vs. \texttt{CVCV}), to create four different sets of novel character sequences. Then we replaced the context-controlled lexical items with the sampled sequences to create four modified datasets (see Table~\ref{table:char-sampling} for examples).

While these sampled sequences are less likely to occur in the pretraining data than real words like \textit{hedgehog}, it is not guaranteed that they are completely absent. As an additional verification step, we searched through the C4 corpus\footnote{\url{https://c4-search.apps.allenai.org/}} to ensure that the sampled sequences are absent from the data that the models we tested (the T5 series) were pretrained on.

\subsection{Model and Training}
We used the T5-base model, which was pretrained on 1 trillion tokens of English text from the \href{https://www.tensorflow.org/datasets/catalog/c4}{C4} corpus. We used the codebase from \citet{orhan2021compositional} that had reported the best pretrained model performance at the time of the experiment (around 83\% generalization accuracy). We finetuned T5-base for a large fixed number of steps (300K, $\sim$398 epochs) without early stopping, following the observation of \citet{csordas2021devil} that generalization may continue to improve even when development set performance saturates.\footnote{Note that, for fair evaluation, we did not tune the number of steps based on generalization set performance. We selected a sufficiently large number of steps that led to near-perfect ($\geq$98\%) development set accuracy in most model variations we tested, as well as 100\% accuracy on the exposure examples in the training set. Learning the exposure examples in the training set like \ref{ex:exposure} that contains the context-controlled lexical items is critical, because it is a precondition to expect any generalization involving those items.} Other hyperparameters were kept equal to \citet{orhan2021compositional} (batch size=32, AdamW optimizer, linear scheduling) except for the learning rate that was tuned based on exposure example accuracy and development set performance ($lr \in \{1 \times 10^{-3}, 1.5 \times 10^{-5}\}$). We finetuned the model 5 times varying the random seed. Each finetuning run took around 48 hours on a single RTX8000 GPU including development set evaluation at every 5000 steps.

\begin{table*}[t]
    \centering
    \resizebox{1.7\columnwidth}{!}{
    \begin{tabular}{cccccc}
    \toprule
           Embedding init. & \colorbox{beaublue}{\underline{\textbf{Gen.}}} & Test-ID &  Gen. (Lex. only) & Test-Lex & Training steps \\ \midrule\midrule
           \texttt{rand} & 0.323 \small{($\pm$ 0.060)} & 0.998 & 0.368 \small{($\pm$ 0.071)} & 0.793 \small{($\pm$ 0.033)} & 300K \\
           \texttt{avg} & 0.060 & 0.999 & 0.070 & 0.379 \small{($\pm$ 0.013)} & 300K \\
           Unused embeddings & 0.059 & 0.999 & 0.068 & 0.404 \small{($\pm$ 0.024)} & 300K \\\midrule
           No modification & 0.833 & 0.998 & 0.963 & 0.973 & 60K \\
           \bottomrule
    \vspace{0.01cm}
    \end{tabular}
    }
    \caption{Generalization accuracy of T5-base with context-controlled lexical items represented by novel embeddings. No modification results are repeated from Table~\ref{table:char-sampling}. Standard deviations are shown if greater than 0.01.}
    \label{fig:exp1-results}
\end{table*}

\paragraph{Tokenization.} We used the Huggingface implementation of the \href{https://huggingface.co/transformers/model_doc/t5.html\#t5tokenizer}{T5 tokenizer}, which is based on SentencePiece \citep{kudo2018sentencepiece}. Therefore, the character sequences replacing the context-controlled lexical items were tokenized into subword tokens, which include both single- and multi-character tokens.

\subsection{Results}

The results are presented in Table~\ref{table:char-sampling}. The generalization performance of the models using character sequences as context-controlled lexical items was 64--69\%. This is 14--19 percentage points lower than results obtained with the unmodified COGS dataset ($\sim$83\%). This is evidence that uncontrolled lexical exposure discussed in Section~\ref{sec:issue} does indeed lead to an overestimation of generalization performance. Interestingly, the performance across different lengths and sampling strategies was similar. This shows that the models remained robust to lexical items that deviate from typical lexical items of English, successfully learning to treat each as a coherent unit.

\section{Experiment 2: Novel Embeddings as Context-controlled Lexical Items}
\label{exp2:novel-embeddings}
We repeated the evaluation using the second setup proposed in Section~\ref{sec:modification} with novel embeddings added directly to the model's vocabulary as context-controlled lexical items.

\subsection{Model and Training}
As before, we finetuned the pretrained T5-base model on the modified training set of COGS, but this time added the tokens that replaced the context-controlled lexical items to the model vocabulary. The hyperparameters were kept the same except for the learning rate that was tuned based on exposure example accuracy and development set performance ($lr = 1.5 \times 10^{-5}$). Finetuning was run 5 times with different random seeds, and each finetuning run took around 48 hours on a single RTX8000 GPU including intermediate development set evaluation at every 5000 steps.

\paragraph{New vocabulary.} We added new embeddings to the model vocabulary before the finetuning step. Each unique context-controlled lexical item was first replaced with a special token [$w_n$] ($|n|$=21), each of which was assigned a new embedding. We tested three initialization schemes for these new embeddings: the default random normal initialization of the \href{https://github.com/huggingface/transformers/blob/main/src/transformers/models/t5/modeling_t5.py}{Huggingface T5} (\texttt{random}), the average of existing embeddings (\texttt{avg}) with noise (suggested by \citet{hewitt2021initializing} as a way to alleviate the divergence of the novel embeddings from existing pretrained embeddings), and unused embeddings in the embedding layer of the model.\footnote{Leftover embeddings that were never used during training: \url{https://github.com/huggingface/transformers/issues/4875}.}

\paragraph{Tokenization.} We used the same tokenizer as in Experiment 1 except for the newly added tokens. We note that T5's tokenizer treats whitespaces as characters rather than tokenization boundaries, and there exist unexpected decoding behaviors concerning whitespaces and added tokens in the version of the tokenizer we used. We log the details and potential issues in Appendix~\ref{app:tokenization} for reference in future work. 

\subsection{Results}
\label{exp2:results}

The results (Table~\ref{fig:exp1-results}) show that the compositional generalization accuracy of the pretrained models was very poor under this setup of using novel embeddings. Both average and unused embeddings averaged around 6\% generalization accuracy. Random initialization yielded the best results, but still quite poor at around 32\%. This sets a more dramatic lower bound to the compositional generalization performance of pretrained T5 models, indicating
about 51 percentage point overestimation compared to the setting in which the original dataset was used without modification ($\sim$83\%). This low performance contrasts with models trained using novel character sequences as context-controlled lexical items ($\sim$68\%, Table~\ref{table:char-sampling}). This large variation in generalization across different evaluation setups can only be attributed to how the context-controlled lexical items are embedded, since this is the only difference between the evaluation setups.

\begin{table*}[h]
    \centering
    \resizebox{2\columnwidth}{!}{
    \begin{tabular}{ccccccc}
    \toprule    
           \# tokens in pretraining data & \colorbox{beaublue}{\underline{\textbf{Gen.}}} & \colorbox{Apricot}{Test-ID} & Gen. (Lex. only) & \colorbox{YellowGreen}{Test-Lex} & $|$Test-Lex$-$Gen. Lex$|$ & Data Source \\ \midrule
           0 (No pretraining) & 0.749 \small{($\pm$ 0.026)} & 0.994 & 0.874 \small{($\pm$ 0.030)} & 0.902 \small{($\pm$ 0.024)} & 0.028 & -\\
           1M & 0.678 \small{($\pm$ 0.069)} & 0.994 & 0.791 \small{($\pm$ 0.080)} & 0.834 \small{($\pm$ 0.064)} & 0.043 & Wikipedia \\
           5M & 0.602 \small{($\pm$ 0.045)} & 0.991 & 0.703 \small{($\pm$ 0.053)} & 0.727 \small{($\pm$ 0.035)} & 0.025 & Wikipedia \\
           25M & 0.538 \small{($\pm$ 0.033)} & 0.985 & 0.628 \small{($\pm$ 0.038)} & 0.652 \small{($\pm$ 0.069)} & 0.024 & Wikipedia \\
           50M & 0.516 \small{($\pm$ 0.027)} & 0.989 & 0.602 \small{($\pm$ 0.031)} & 0.686 \small{($\pm$ 0.042)} & 0.084 & Wikipedia\\
           100M & 0.787 \small{($\pm$ 0.003)} & 0.999 & 0.918 \small{($\pm$ 0.003)} & 0.942 \small{($\pm$ 0.015)} & 0.024 & Wikipedia\\
           1B & 0.722 \small{($\pm$ 0.036)} & 0.999 & 0.842 \small{($\pm$ 0.042)} & 0.883 \small{($\pm$ 0.029)} & 0.041 & Wikipedia\\
           1T (Full T5-small) & 0.279 \small{($\pm$ 0.026)} & 0.999 & 0.326 \small{($\pm$ 0.030)} & 0.802 \small{($\pm$ 0.070)} & 0.478 & C4\\
           \bottomrule
    \vspace{0.01cm}
    \end{tabular}
    }
    \includegraphics[width=0.75\textwidth]{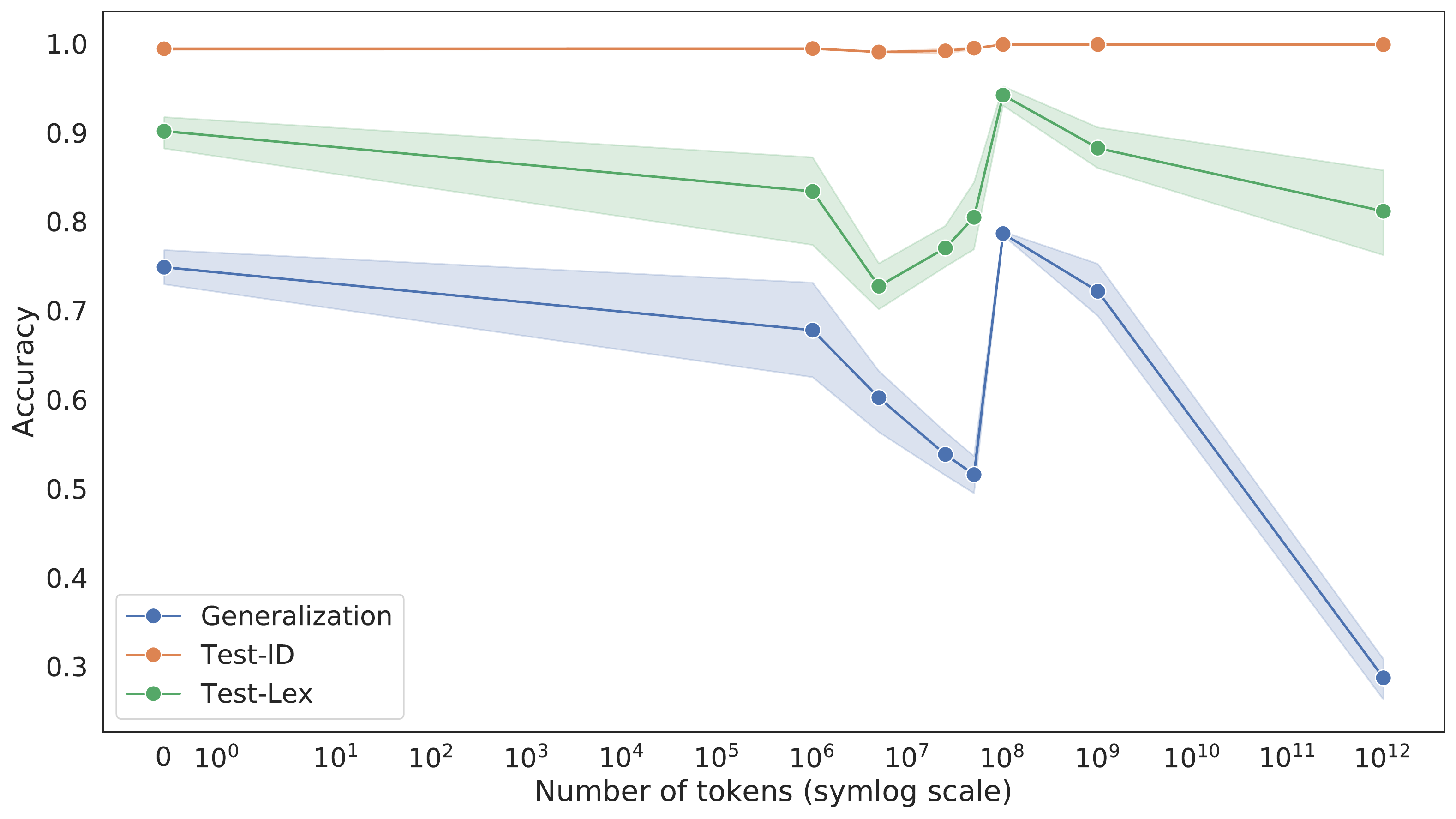}
    \caption{Generalization accuracy of T5-small models pretrained on different amounts of data, with context-controlled lexical items represented by randomly initialized novel embeddings. Standard deviations are shown if greater than 0.01. The $x$-axis shows the number of tokens in symmetrical log scale to include 0 in the plot.}
    \label{fig:exp2-lm-results}
\end{table*}

\paragraph{Are the models simply incapable of producing novel tokens?} One possibility that can lead to the low generalization performance of pretrained models is if the novel tokens added to the vocabulary are never produced because models consistently assign them low probabilities compared to existing tokens. In every experiment, we ensured that the models perfectly learned the exposure examples that contain the novel tokens, as mentioned in Footnote 5. This means at least for the training examples, the models were capable of learning and producing the novel tokens without issue. Furthermore, the models generally had no problem with producing the novel tokens even outside of these particular training examples. In fact, $\sim$97\% of the model predictions for lexical generalization contained at least one novel token, as they should, although the prediction itself was still incorrect. Therefore, the low performance of these models cannot be attributed to a total incapacity to produce newly added tokens.

\paragraph{Poor generalization cannot be reduced to lexical difficulty.} The performance on \textsc{Test-Lex} (Section~\ref{subsec:id-test-novel-lex}) provides more insights into the source of the large performance degradation. The models did struggle more on \textsc{Test-Lex} (38--79\%) than on the original \textsc{Test-ID} which contains no context-controlled items ($\sim$99\%). This discrepancy within in-distribution tests shows that rare lexical items are challenging, which likely accounts for some portion of the generalization degradation we observe. However, the target compositional generalization set performance is impacted over and above this degradation due to lexical difficulty: there is a further gap between \textsc{Test-Lex} and \mbox{\textsc{Gen.\ (Lex.\ only)}} (38--79\% vs. 6--32\%). These two evaluation sets are equivalent in that they contain context-controlled items, but differ in terms of their contextual difficulty. Therefore, the lower accuracy in the generalization set must derive from contextual difficulty rather than lexical difficulty.

\section{Experiment 2+: Effect of Pretraining Corpus Size on Generalization}
Experiment 2 showed that the generalization performance of T5 was extremely poor when novel embeddings were used to represent context-controlled lexical items. In this follow-up experiment, we investigate if we can attribute the low generalization performance specifically to the amount of data the model has been exposed to. We approach this question by comparing multiple models of the same architecture that vary only in the amount of pretraining data, including a model without any pretraining.

\subsection{Model and Training}
\label{subsec:model-training}
We used the T5-small model for this experiment.\footnote{The choice of T5-small over other larger variants such as T5-base from the previous experiments is due to resource constraints. Note that the difference in generalization performance between fully pretrained T5-small and T5-base under the novel embeddings setup is marginal (7.6\% vs. 5.9\%).} We first randomly initialized the T5-small model and pretrained it on varying amounts of data using the span corruption objective: 0 (i.e., not pretrained), 1M, 5M, 25M, 50M, 100M, and 1B tokens. We used 10\% of the datasets as development sets to determine early stopping points with a patience of 5. Then, we finetuned each model on COGS, using the novel embeddings setup in Experiment 2. We used \href{https://www.tensorflow.org/datasets/catalog/wikipedia}{English Wikipedia} instead of C4 for pretraining due to resource limitations in running the preprocessing pipeline of C4.

Finetuning was run 5 times for each model using different random seeds for 500K steps---the number of steps sufficient for the models to learn the exposure examples perfectly and achieve near-perfect in-distribution development set accuracy. The learning rate was tuned based on development set performance, and other hyperparameters were the same as Experiment 2. Finetuning took around 30 hours on a single RTX8000 GPU including intermediate development set evaluation at every 5000 steps. We used random initialization for the novel embeddings, since \texttt{rand} and \texttt{avg} did not differ meaningfully in Experiment 2. 

\subsection{Results}
\vspace{0.2cm}
Table~\ref{fig:exp2-lm-results} shows the generalization accuracy of T5-small models pretrained with varying amounts of data.\footnote{The randomly initialized T5-small generalized better than T5-base, which replicates the finding in \citet{orhan2021compositional} that larger models are harder to train from scratch on COGS.} First of all, a fully pretrained model performed much worse than a randomly initialized model of the same architecture (28\% vs.\ 75\%), demonstrating a negative impact of pretraining under the novel embeddings setup. Overall, generalization performance is negatively correlated with the amount of pretraining data (Spearman's $\rho$ = $-0.29$, $p=.07$). Importantly, the gap between \textsc{Test-Lex} and the lexical portion of the generalization set ($|$Test-Lex$-$Gen.\ Lex$|$) increased with the amount of training data (\textit{r} = $0.45$, $p<.01$). This demonstrates that the capacity to handle contextual novelty through composition is damaged by pretraining under the novel embeddings setup, over and above the general adverse effect on the processing of novel tokens, as discussed in Section~\ref{exp2:results}. This finding illustrates an interesting case of inverse scaling\footnote{Cases in which task performance gets worse as parameters, compute, and/or data size increase: \url{https://github.com/inverse-scaling/prize}.} in the T5 series.

\section{Discussion}
\label{sec:the-question}
\begin{figure}[t]
\centering
    \includegraphics[width=0.49\textwidth]{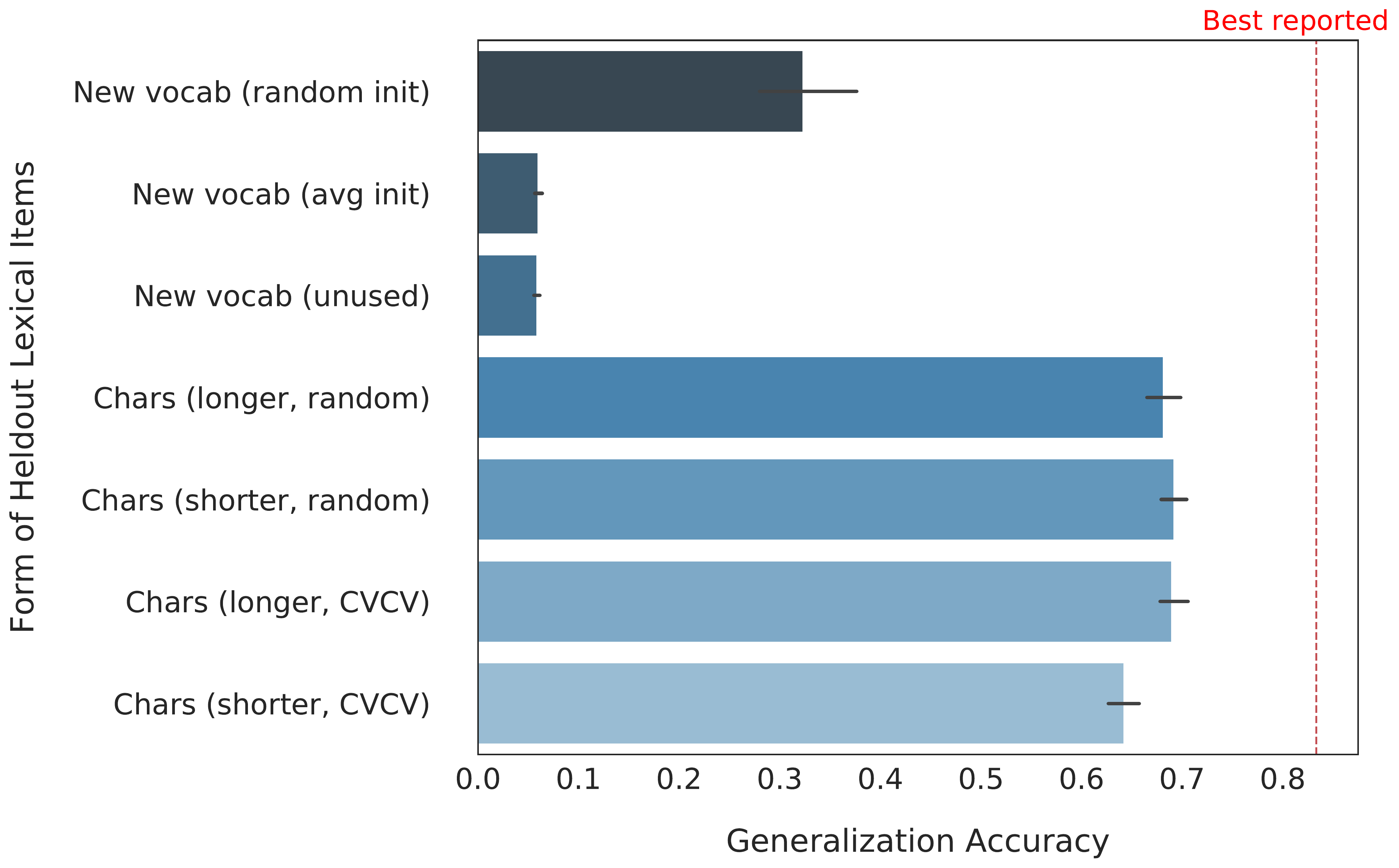}
    \caption{Highly variable generalization performance of T5-base under different modifications proposed in this paper. Best reported performance using T5-base from \citet{orhan2021compositional} is marked with a red dotted line. \textbf{Overestimation} refers to the difference between this red dotted line and the blue bars.}
    \label{fig:t5-various-lexical-reps}
\end{figure}

We have shown through a series of experiments that the compositional generalization performance reported in the literature using pretrained models, in particular T5, is likely overestimated. In addition to this general conclusion, one surprising finding is that the choice of context-controlled lexical items had a large impact on the generalization outcomes (Figure~\ref{fig:t5-various-lexical-reps}), despite the models being extensively finetuned. Using character sequences (Experiment 1) and novel embeddings (Experiment 2) led to dramatically different generalization performance, even though the only difference across these experiment setups was the form of the context-controlled items.

This sensitivity to lexical initialization when all exposure examples are perfectly learned would not arise in models that generalize systematically---if they have learned to assign a correct meaning representation to \textit{John saw Mary}, then \textit{Mary saw John} should also be correctly mapped to the correct meaning representation. In a truly systematic model that operates on the basis of ``operations on symbols'' or ``algebraic manipulation'' \citep{newell1980physical,pylyshyn1980computation,fodor1988connectionism,marcus2001algebraic}, we would observe similar degrees of success across all variations of the experiments in this paper. However, this was not the case with the models we tested. Is this a problem?

Arguably, novel character sequences cover a large portion of expected downstream instances in which lexical generalization will be required from models such as T5. While we observed a nontrivial amount of overestimation under this setup, the performance remained competitive, sometimes even outperforming approaches designed to tackle compositional generalization specifically (e.g., \citealt{conklin2021meta}). The models were also robust to the length and sampling strategy of the character sequences, suggesting that they can treat arbitrary character sequences as a coherent unit.

Even if we consider human generalization capacity as a reference point, the experimental conditions under which empirical evidence for human generalization have been obtained seem similar to this setup. In human subject experiments that test similar types of generalization, nonce words like \textit{gorp}, \textit{mib}, and \textit{pilk} (which obey the phonotactics of the target language, here English) often play the role of context-controlled lexical items (e.g., \citealt{olguin1993twenty,kline2014syntactic}). This seems most comparable to the shorter CVCV sampling case in Experiment 1, in terms of the properties of the nonce words. It is unclear as of now what the analogous setup to Experiment 2 (novel embeddings) would be in human learners, but this is an interesting question for future work investigating the human capacity for compositional generalization.

In light of this discussion, we believe that the choice of evaluation should be informed by the research question one wishes to address. If the question is about robustness to average out-of-vocabulary encounters in the wild, the character sequence substitution approach seems sufficient. However, if one's downstream use case involves training novel embeddings (e.g., novel entities, ontological changes), robustness to novel embeddings would be critical. Finally, the goal may be scientific: investigating whether a certain neural network underlyingly implements a classical symbolic system (in the sense of \citealt{mclaughlin1993connectionism} and others) by probing for generalization that is invariant to the choice of lexical initialization. Here, one may consider a wider range of experiments, including both methods we proposed and possibly others, and test whether generalization is stable.

We note that we do not engage in broader discussions about whether generalization in human learners is actually achieved on the basis of abstract symbolic manipulation, whether this kind of capacity is a precondition to intelligence, or whether this \textit{ought} to be the kind of model that we should be building. Our points are as follows: (1) it is good practice to spell out what capacity one wants to probe through benchmarks, and to adopt a setup that aligns with the research question, and (2) in any case, directly evaluating pretrained models on compositional generalization tests that depend on lexical control without implementing adequate control measures is misleading.\footnote{One may argue that since the compositional generalization task is distinct from the pretraining task or other possible auxiliary tasks, it follows that maintaining the intended distributional gap between training and generalization at finetuning time suffices. Our view is that enforcing distributional control at finetuning time only is addressing a different research question, namely whether models can adapt to a specific finetuning task under distribution shift of certain lexical items. The original tests intend to evaluate generalizations that rely on the underlying linguistic system  inherently connecting certain expressions (e.g., \textit{X saw Y}) to others (e.g., \textit{Y saw X}), so as to allow for the application of compositional rules even in the \textit{absence} of observing the relevant expressions directly (e.g., knowing what \textit{X saw Y} means entails being able to generalize to \textit{Y saw X} despite never having encountered this expression). This question cannot be correctly posed if any explicit evidence about the target generalizations, purely distributional (in the language modeling sense) or otherwise, is provided in addition to the training data of the benchmark tests. This is an argument based on principle, but this work can also be viewed as empirically testing whether uncontrolled lexical exposure does in fact have a substantial impact on models' generalization behavior.} We once again invoke the analogy to human nonce word experiments: imagine that the famous wug test \citep{berko1958child} was in the form of \textit{This is a slug. Now there are two of them. There are two \_\_.}, with the real word \textit{slug} in place of \textit{wug}. Even if a subject produces the expected ending /z/, this result cannot serve as evidence for the existence of an abstract pluralization rule, because prior observations of \textit{slugs} could have been retrieved from memory. The same analogy applies to using real words as context-controlled lexical items in compositional generalization benchmarks when pretrained models are being tested.

\section{Related Work}
Methodologically, this work is closest to approaches that make use of novel embeddings to evaluate the generalization capacity of pretrained models \citep{kim2021testing,petty2022language}. More broadly, this work has connections to discussions about the implications of lexical representation and tokenization in Natural Language Processing \citep[\textit{i.a.}]{domingo2018much,mielke2021between,xue2022byt5}.

Regarding compositional generalization, our findings potentially impact the interpretation of a large body of existing work in this domain that uses pretrained models \citep[\textit{i.a.}]{furrer2020compositional,tay2021pretrained,shaw2021compositional,orhan2021compositional,qiu2021improving,zhu2021learning,herzig2021unlocking,qiu2022evaluating,zheng2022disentangled,drozdov2022compositional}, where the benefit of pretraining is most prominent for lexical generalization.

In general, lexical generalization is known to be less challenging for contemporary neural networks (a stronger statement from \citealt{weienhorn2022compositional}: ``lexical generalization is essentially a solved problem for seq2seq models''). There are several almost-perfect solutions for lexical generalization that do not rely on pretraining \citep{bergen2021systematic,akyurek2021lexicon}, the solutions sometimes being as simple as changing the training configurations of vanilla seq2seq models \citep{csordas2021devil}. In this context, the current work highlights a new difficulty concerning lexical generalization: reconciling pretraining and robustness to the choice of lexical initialization.

\section{Conclusion}
Compositional generalization benchmarks such as SCAN and COGS are often used to evaluate pretrained models. We have shown that the interpretation of such experiments can be complicated by the fact that pretrained models likely violate the control for lexical exposure that these benchmarks depend on to measure generalization. We have proposed modifications based on lexical substitution to remedy this issue and presented empirical results on how these modifications affect the generalization outcomes, using the COGS dataset as a testbed. The results indicate that the generalization performance of the T5 model drops significantly compared to previously reported results (83\% $\rightarrow$ 6--68\%) when trained on the version of the dataset with the proposed modifications. This shows that there is a measurable effect of uncontrolled lexical exposure. When evaluated without adequate control measures, pretrained models likely have observed the key lexical items during pretraining many times, and possibly also as parts of constructions that these lexical items should be withheld from, which leads to overestimated generalization performance.

The degree of performance degradation greatly varied depending on the lexical substitution strategy adopted in the two proposed control setups. With character sequences, the performance gap with was around 14--19 percentage points, whereas with novel embeddings, the gap was as large as 51 points. Furthermore, we found that in the novel embeddings case, randomly initialized models substantially outperformed pretrained models. This harmful effect of pretraining contrasts with previously reported benefits of pretraining for compositional generalization (e.g., \citealt{tay2021pretrained,orhan2021compositional}).

How should we interpret this high variance of results across different control methods, and how should we move forward with using compositional generalization benchmarks to evaluate pretrained models? We argue that there is no one-fits-all solution, and the right evaluation depends on one's research question. For example, if what is being evaluated is a truly systematic generalization that does not depend on specific choice of lexical items, the T5 models we tested did not show this kind of a robust capacity. If what is being evaluated is the capacity to generalize in expected use case scenarios covered by subword-based representations, the models we tested showed some degree of success, albeit their generalization performance being significantly lower than what has been previously reported.

\section*{Acknowledgements}
We thank Kyle Rawlins, Sebastian Schuster, Kanishka Misra, members of the Computation and Psycholinguistics Lab, and members of the Human \& Machine Learning Lab for discussions about this project. We thank Emin Orhan for his detailed feedback and suggestions about the experiments in this paper and the pointer to average word embedding initialization. We thank Santiago Onta\~{n}{\'o}n for the results on the larger T5 models in the Appendix. This work was supported by NSF BCS-204122, and in part through the NYU IT High Performance Computing resources, services, and staff expertise.

\bibliography{paper}
\bibliographystyle{acl_natbib}

\appendix
\section{What about Structural Generalization?}
\label{app:structural-generalization}
Even with our proposed modification, the distributional control for \textbf{structural} generalization (generalization to unseen structures, e.g., to deeper degrees of embedded structures) cannot be guaranteed without full structural inspection of the pretraining data. For example, we would have to remove all examples containing 3+ nested prepositional phrases from the pretraining data. Implementing this control is especially challenging under a common scenario where the model's pretraining data is not publicly available or difficult to inspect due to its size (e.g., models trained on all of Wikipedia, models trained on BooksCorpus which is no longer publicly available, models trained on company-internal data). Furthermore, retraining the models on the dataset with target structures removed would pose additional challenges.

In the case of generalization to deeper degrees of embedding, it is reasonably likely that embeddings of depth $\geq$ 6 would not occur in the pretraining data \citep{karlsson2010syntactic}, but this is only speculative. We leave the issue of enforcing distributional control for structural generalization for pretrained models to future work, but note that there is no known meaningful benefit of language modeling pretraining alone for structural generalization even under the uncontrolled setting. Most of the known gains have been lexical; for instance, Table~\ref{table:lex-struct-t5} shows that structural generalization accuracy of T5 finetuned on unmodified COGS is very poor. 

\begin{table}[h]
    \centering
    \resizebox{\columnwidth}{!}{
    \begin{tabular}{cccc}
    \toprule    
           Model & Gen. (all) & Lexical & \colorbox{GreenYellow}{Structural} \\\midrule
           T5-base (rand.) & 0.439 & 0.511 & 0 \\
           T5-base & 0.833 & 0.963 & 0.053 \\\midrule
           T5-large & 0.832 & 0.971 & 0 \\
           T5-xl & 0.711 & 0.829 & 0.001 \\
           T5-xxl & 0.836 & 0.974 & 0.006 \\
           \bottomrule
    \end{tabular}
    }
    \caption{Generalization accuracy of T5 models \underline{without} applying any modification. The larger T5 models are from finetuning the publicly available checkpoints of T5 v1.1, and were run with the help of Santiago Onta\~{n}{\'o}n.}
    \label{table:lex-struct-t5}
\end{table}

\section{Tokenization for Added Vocabulary}
\label{app:tokenization}
Here, we document the issues that we encountered while implementing the vocabulary expansion using the Huggingface version of T5, which potentially causes problems with the exact string match metric because of misligned whitespaces. There are two available tokenizers compatible with this implementation, T5TokenizerFast and T5Tokenizer. Our goal is to add tokens of the form `[$w_0$]' to the model. Since whitespace is considered a character, `[$w_0$]' and ` [$w_0$]' are considered to be different tokens. To achieve the intended behavior in the model we used, the following needs to be done.

\vspace{0.05cm}
\noindent \textbf{T5TokenizerFast}: either (1) both whitespace prepended ( [$w_0$]) and bare ([$w_0$]) versions of the token should be added to the tokenizer, IN THIS ORDER, or (2) when the context-controlled lexical items are replaced at the dataset level, we can replace the sequence-initial context-controlled tokens with the whitespace prepended version and add only this version to the tokenizer. We provide more detailed descriptions of the possible scenarios:

\begin{enumerate}
  \setlength{\itemsep}{1pt}
  \setlength{\parskip}{1pt}
    \item If only the bare version is added, the whitespace before the novel token will be dropped at decoding time, leading to erroneous spacing sequence-medially.
    \item If only the whitespace prepended version is added, sequence-initial novel tokens will not be tokenized as a single token (e.g., [$w_0$] -> `[', `w', `\_',  `0', `]').
    \item If both are added but in reverse order (bare then whitespace), sequence-medial novel tokens will be tokenized as the bare version, and the whitespace originally preceding this token will be lost at decoding time.
    \item If sequence-initial tokens are replaced with the whitespace prepended version in the dataset itself, we will get the desired behavior by just adding the whitespace prepended version to the tokenizer.
\end{enumerate}

\noindent \textbf{T5Tokenizer}: just adding the bare version or both whitespace prepended \& bare versions in any order works, but just adding the bare version has caveats. Just adding the whitespace prepended version must be avoided.
\begin{enumerate}
  \setlength{\itemsep}{1pt}
  \setlength{\parskip}{1pt}
    \item Even if only the bare version is added, the whitespace before the novel token that occurs sequence-medially will not be lost at decoding time. Hence, this will not cause issues with exact match evaluation. However, the actual tokenization will still not have the prepended whitespace, which is different from how typical sequence-medial tokens are treated in T5. So perhaps, adding both versions is a better approach.
    \item If only the whitespace prepended version is added, sequence-initial novel token will not be tokenized as a single token as in the case of T5TokenizerFast.
\end{enumerate}

\noindent We used T5TokenizerFast with the 4th option listed above, but also sanity checked that there is no substantial performance gap between valid options. Also note that although whitespaces in T5 tokenizers are represented by `\textbackslash u2581', in some versions of the tokenizer, when adding a whitespace prepended token to the tokenizer, only ` ' instead of `\textbackslash u2581' will lead to intended behaviors. In such versions, if `\textbackslash u2581' is used, the tokenizer will not correctly tokenize the novel tokens and they will be subword tokenized.

\end{document}